%% file: aied2021.tex
\documentclass[runningheads]{llncs}

\usepackage{amssymb,amsmath,amsfonts}
\usepackage{graphicx}
\usepackage{url}
\usepackage{color}
\usepackage{amsfonts}
\usepackage{blindtext}
\usepackage{mathrsfs}
\usepackage{bbm}
\usepackage{url}
\usepackage{paralist}
\usepackage[english]{babel}
\usepackage{booktabs}
\usepackage{multicol}
\usepackage{multirow}
\usepackage{subfigure}
\usepackage{amsmath}
\usepackage{graphicx}
\begin{document}
\title{Solving ESL Sentence Completion Questions via Pre-trained Neural Language Models}
\titlerunning{Solving ESL Sentence Completion Questions}
%
\author{Qiongqiong Liu\inst{1}, Tianqiao Liu\inst{1}, Jiafu Zhao \inst{1}, Qiang Fang \inst{1}, Wenbiao Ding \inst{1}, Zhongqin Wu \inst{1}, Feng Xia \inst{3}, Jiliang Tang \inst{2}, Zitao Liu \inst{1}\thanks{Corresponding Author: Zitao Liu}}


\authorrunning{Q. Liu et al.}
%

\institute{TAL Education Group, Beijing, China \email{\{liuqiongqiong1,liutianqiao,zhaojiafu,fangqiang,dingwenbiao,wuzhongqin, liuzitao\}@tal.com} \and
Data Science and Engineering Lab, Michigan State University, USA
\email{tangjili@msu.edu} \and
Federation University Australia, Australia \\
\email{f.xia@ieee.org} 
}

%
%
\maketitle              
%
\begin{abstract}
\input{abstract}
\keywords{Sentence completion \and Pre-trained language model \and Neural networks.}
\end{abstract}

\section{Introduction}
\label{sec:intro}
\input{intro.tex}

\section{Our Approach}
\label{sec:approach}
\input{approach.tex}

\section{Experiments}
\label{sec:exp}
\input{experiment.tex}

\vspace{-0.2cm}
\section{Conclusion}
\label{sec:conclusion}
\input{conclusion.tex}

\vspace{-0.1cm}
\section*{Acknowledgment}

This work was supported in part by National Key R\&D Program of China, under Grant No. 2020AAA0104500 and in part by Beijing Nova Program (Z201100006820068) from Beijing Municipal Science \& Technology Commission.
%
%
%
\bibliographystyle{splncs04.bst}
\bibliography{aied2021}
\end{document}

%% file: abstract.tex
Sentence completion (SC) questions present a sentence with one or more blanks that need to be filled in, three to five possible words or phrases as options. SC questions are widely used for students learning English as a Second Language (ESL) and building computational approaches to automatically solve such questions is beneficial to language learners. In this work, we propose a neural framework to solve SC questions in English examinations by utilizing pre-trained language models. We conduct extensive experiments on a real-world K-12 ESL SC question dataset and the results demonstrate the superiority of our model in terms of prediction accuracy. Furthermore, we run precision-recall trade-off analysis to discuss the practical issues when deploying it in real-life scenarios. To encourage reproducible results, we make our code publicly available at \url{https://github.com/AIED2021/ESL-SentenceCompletion}.

%% file: intro.tex
Sentence completion (SC) questions present a sentence with one or more blanks that need to be filled in. Three to five possible words (or short phrases) are given as options for each blank and only one of the options yields to a reasonable sentence. SC questions have been proven a necessary source of evaluation data for investigating and diagnosing the situations that the English as a Second Language (ESL) learners grasp the essential language knowledge \cite{franke1960reform,madaus1991effects,davey2007university,beinborn2015candidate,zweig2012computational}. An example of SC question is shown in Table \ref{tab:example}.

\begin{table}[!bhpt]
\caption{An illustrative example of SC questions.} \vspace{-0.75cm}
\begin{center}
\begin{tabular}{@{}ll@{}} \toprule
\label{tab:example}
 & — That T-shirt with Yao Ming's picture on it $\underline{\quad\quad}$ belong to John. He likes  him a lot. \\ 
 & — No, it $\underline{\quad\quad}$ be his. He hates black color. \\
 & (A) can; can't (B) may; needn't (C) must; mustn't (D) must; can't \\  
\bottomrule
\end{tabular}
\end{center}
\vspace{-0.75cm}
\end{table}

In this work, we study computational approaches to automatically solve such ESL SC questions. They are valuable for many reasons: (1) they are able to provide instant feedback to students and help students learn and practice ESL questions anytime anywhere; (2) they provide feasible solutions to evaluate distractors in SC questions and help teachers revise and improve the overall qualities of SC questions; and (3) they shed light on the opposite tasks like automatically generating questions for language proficiency evaluation and provide as many as possible training samples for building effective question-answering systems or intelligent tutoring systems.

Various approaches have been proposed to automatically solve the ESL SC questions. For example, Zweig et al. chose to use a trigram language model (LM) for solving the SC questions in Scholastic Aptitude Test (SAT) where the trigram LM is trained on 1.1B words from newspaper data \cite{zweig2012computational}. Shen et al. proposed a blank LM to iteratively determine which word to place in a blank and whether to insert new blanks, until no blanks need to be filled \cite{shen2020blank}. Donahue et al. trained the LM by using the concatenation of artificially-masked texts and the texts which are masked as input \cite{donahue2020enabling}. 

However, automatically solving ESL SC questions still presents numerous challenges that come from special characteristics of real-world educational scenarios as follows: (1) confusing distractors: the ESL SC questions are created by English teaching professionals and the corresponding distractors are very similar; (2) detailed linguistic knowledge: due to the evaluation propose, SC questions always embed detailed linguistic knowledge including grammar, syntax, and semantics; and (3) arbitrary number of blanks and tokens: the ESL SC questions may have one or more missing blanks to be filled and each of which may require an arbitrary unknown number of tokens.

To overcome the above challenges, we propose to utilize a large-scale neural LM to automatically solve the ESL SC questions in students' real-life scenarios. Our approach is based on the standard Transformer-based neural machine translation architecture and utilizes a denoising autoencoder for pre-training sequence-to-sequence models. Our approach shows a powerful generalization capability for automatically solving ESL SC questions of various types from real-world scenarios. Experiments conducted on a real-world online education dataset demonstrate the superiority of our proposed framework compared with competitive baseline models.

%% file: approach.tex
The SC question is composed of (1) a question, i.e., $\mathbf{q}$, formed in natural language with one or more blanks, and (2) $m$ candidate options, i.e., $\mathbf{o}_1, \cdots, \mathbf{o}_m$. Solving the SC question is to find the option that leads to the highest correct probability after completing the to-be-filled sentence with the selected option, i.e., $\arg\max_{i = 1, \cdots, m} \mbox{Pr}(\mathbf{o}_i | \mathbf{q})$.

In this work, we first fill candidate options into the corresponding blanks to get complete sentences. Then we treat sentences that contain the correct options as positive examples and the rest as negative examples. After that, we build a neural LM model to extract the semantically meaningful information within each sentence and make final SC question predictions via a multilayer perceptron (MLP).

We choose to use a denoising autoencoder for pretraining sequence-to-sequence models, i.e., BART, \cite{lewis2019bart} as our neural LM model. BART adapts standard Transformer \cite{vaswani2017attention} as its backbone model and is pre-trained to map corrupted document to their original. We apply pre-trained BART model to our SC questions task with simple modifications on the output layers and loss function. Specifically, given a complete sentence $\mathbf{q} = (w_1, w_2, \cdots, w_n)$, we first convert it into token embeddings $\mathbf{E} = (\mathbf{e}_1, \mathbf{e}_2, \cdots, \mathbf{e}_n)$, where $\mathbf{E} \in \mathbb{R}^{n \times d}$, and $d$ is the embedding size. Then we pass $\mathbf{E}$ through multiple Transformer encoder layers to obtain the contextualized token representations $\mathbf{H} = (\mathbf{h}_1, \mathbf{h}_2, \cdots, \mathbf{h}_n)$. The input of the decoder is the same as the encoder and we pass $\mathbf{E}$ to a stack of Transformer decoder layers. Different from the encoder, masked self-attention is applied to ensure that the predictions can depend only on the information at prior positions in the decoder. Additionally decoder performs cross-attention over the final hidden representations of the encoder, i.e., $\mathbf{H}$. Finally, we obtain the final hidden states $(\mathbf{t}_1, \mathbf{t}_2, \cdots, \mathbf{t}_n)$, where $\mathbf{t}_i \in \mathbb{R}^{d \times 1}$. We utilize the final hidden state $\mathbf{t}_n$ as the aggregated sentence representation. We introduce two additional fully-connected layers to perform the binary classification task, i.e., $\mathbf{x} = \mathrm{softmax}(\mathbf{W}_1 \mathrm{tanh}(\mathbf{W}_0 \mathbf{t}_n + \mathbf{b}_0) + \mathbf{b}_1))$, where $\mathbf{W}_0 \in \mathbb{R}^{1024 \times d}$, $\mathbf{b}_0 \in \mathbb{R}^ {1024}$, $\mathbf{W}_1 \in \mathbb{R}^{2 \times 1024}$ and $\mathbf{b}_1 \in \mathbb{R}^{2}$. The first entry of $\mathbf{x}$ gives the probability of wrong option while the second entry gives right option probability. The objective is to minimize the cross entropy of the right or wrong option labels.

%% file: experiment.tex
We collect real-world K-12 English SC exam questions from a thirty-party educational company. After data cleaning and random shuffling, we end up with 250,918 and 48,686 SC questions as our training and testing datasets. Due to the fact that the difficulty of a particular SC question heavily depends on the number of to-be-filled blanks and the number of tokens in the candidate options. Therefore, we divide the SC questions into the following four categories: C1: one-blank and one-token; C2: one-blank and many-token; C3: many-blank and one-token; and C4: many-blank and many-token. Specifically, we have 114,547, 138,392, 28,738 and 17,927 SC questions in each category.

We carefully choose the following state-of-the-art pre-trained LM approaches as our baselines (1) BERT \cite{devlin2019bert}: a pre-trained natural language understanding model with transformer encoder blocks; (2) XLNet \cite{yang2019xlnet}: an autoregressive based pre-training method with transformer decoder blocks; (3) ELECTRA \cite{clark2020electra}: a more sample-efficient pre-training framework which adapts a generator to perform the textualized masked language modeling task and a discriminator to perform token-level ``real-fake" binary classification task; (4) RoBERTa \cite{liu2019roberta}: improves BERT by replacing static masking with dynamic masking, pre-training more epochs with larger batch size, and removing the next sentence prediction task.

\vspace{-0.3cm}
\subsection{Results}
\label{sec:results}

As we can see from Table \ref{tab:MResults}, our model outperforms all other methods in terms of prediction accuracy on all SC question categories. Specifically, when comparing the prediction performance of all the methods on C1 to C2, C3 and C4, we can see that the increase of either the number of blanks or the length of options does not hurt the accuracy of ESL SC question solvers. The pre-trained large-scaled LMs are very robust and insensitive to SC questions in different categories. 

\begin{table}
\footnotesize
\vspace{-0.5cm}
    \caption{Results on different categories of SC question datasets in terms of accuracy.}
        \label{tab:MResults}\vspace{-0.2cm}
    \begin{center}  
    \begin{tabular}{lccccc}
    \toprule
              & C1              & C2              & C3              & C4                         \\ \midrule
BERT          & 0.8840          & 0.8894          & 0.9221          & 0.9166                    \\ 
XLNet         & 0.9128          & 0.9165          & 0.9290          & 0.9264                    \\ 
ELECTRA       & 0.9212          & 0.9186          & 0.9346          & 0.9236                    \\ 
RoBERTa       & 0.9171          & 0.9321          & 0.9380          & 0.9304                    \\ 
\textbf{BART} & \textbf{0.9381} & \textbf{0.9428} & \textbf{0.9475} & \textbf{0.9445}   \\ \bottomrule
    \end{tabular}
    \end{center}
    \vspace{-1cm}
\end{table}

Furthermore, we conduct a precision-recall trade-off analysis on the results. When deploying the model in practice, a wrong answer may give bad guidance to students. In order to reduce such problem, we may refuse to solve some difficult questions and improve the precision of more solvable questions. We set a threshold to the correct probability of the model's selected option and accept the above-the-threshold questions as our solvable questions. The recall is computed as (the number of solvable questions)/(the number of all test questions), and the precision is calculated as (the number of both solvable and correct-answered questions)/(the number of solvable questions). Finally, we find that when the threshold is 0.95, the precision reaches 97.22\% and the recall is 88.17\% which can be used in practice.

%% file: conclusion.tex

In this paper, we present a neural framework for automatically solving the ESL sentence completion questions. Experimental results based on the real-world English examinations indicate that our proposed model works well in different kinds of sentence completion questions. Furthermore, we conduct fine-grained performance analysis on ESL SC questions from different categories and a trade-off analysis between precision and recall, which reveals insights of applying the proposed approach in the real-world production system.